\documentclass[smallextended]{svjour3}       
 \usepackage{graphicx}
\usepackage[cmex10]{amsmath}

\usepackage{amsthm}

\usepackage[tight,footnotesize]{subfigure}

\begin{document}

\title{Nonlinear behavior of memristive devices and their impact on learning algorithms like STDP in memristive neural networks}

\author{Farnood Merrikh-Bayat         \and
        Saeed Bagheri Shouraki \and Iman Esmaili Paeen Afrakoti 
}


\institute{Farnood Merrikh-Bayat \at
              Department of Electrical Engineering, Sharif University of Technology, Tehran, Iran\\
              \email{f\_merrikhbayat@ee.sharif.edu}           
           \and
           Saeed Bagheri Shouraki\at
             Department of Electrical Engineering, Sharif University of Technology, Tehran, Iran\\
              \email{bagheri-s@sharif.edu}           
           \and
           Iman Esmaili Paeen Afrakoti\at
             Department of Electrical Engineering, Sharif University of Technology, Tehran, Iran\\
              \email{i\_esmaili@ee.sharif.edu}           
}

\date{Received: date / Accepted: date}

\maketitle


\begin{abstract}
It is now widely accepted that memristive devices are perfect
candidates for the emulation of biological synapses in neuromorphic
systems. This is mainly because of the fact that like the strength
of synapse, memristance of the memristive device can be tuned
actively ({\it e.g.}, by the application of voltage or current). In
addition, it is also possible to fabricate very high density of
memristive devices (comparable to the number of synapses in real
biological system) through the nano-crossbar structures. However, in
this paper we will show that there are some problems associated with
memristive synapses (memristive devices which are playing the role
of biological synapses). For example, we show that the variation
rate of the memristance of memristive device depends completely on
the current memristance of the device and therefore it can change
significantly with time during the learning phase. This phenomenon
can degrade the performance of learning methods like Spike
Timing-Dependent Plasticity (STDP) and cause the corresponding
neuromorphic systems to become unstable. Finally, at the end of this
paper, we illustrate that using two serially connected memristive
devices with different polarities as a synapse can somewhat fix the
aforementioned problem.
\end{abstract}


%

\section{introduction}

Publication of a paper \cite{williams} in Nature by HP labs in May
1, 2008, which announced the first experimental realization of the
memristive device whose existence was predicted in 1971 by Leon Chua
\cite{Chua} has caused an extraordinary increased interest in this
passive circuit element. A memristive device is a device that, like
a resistor, opposes the passage of current. But memristive devices
also have a memory. The memristance of a memristive device at any
moment depends on the history of the applied voltage, so its
behavior can be used to recall past voltages. One of the widely
accepted application of memristive devices is in the hardware
implementation of synapses in neuromorphic systems (for example see
\cite{lu,Cantley20,pershin20,Snider20}). Using memristive devices as
synapses in neuromorphic systems can offer both high connectivity
and high density (through the memristive crossbar structure) which
are necessary for efficient computing. However, in this paper we
will show that although memristive devices behave so similar to
biological synapses, they have an unwanted property that may affect
the efficiency of the hardware implementation of neuromorphic
systems considerably.

The rest of this paper is organized as follows.  In Section
\ref{Sec1} we show that memristive devices have this property that
their memristance change due to the applied voltage or current with
a rate which is proportional to their initial memristances. Moreover, we show
how this phenomenon can affect the working procedure of one of the
well-known learning methods in neuromorphic systems {\it i.e.} STDP.
The illustration of how this aforementioned problem can cause
neuromorphic systems to become unstable  is presented in
Section \ref{Sec2}. Our proposed method to relax this problem is
explained in Section \ref{Sec3}, and finally, Section \ref{Sec4}
concludes the paper.

\section{Some Problems behind using memristive devices as synapse}
\label{Sec1}

\begin{figure}[!t]
\centering 
{
\includegraphics[width=3.1in,height=2.2in]{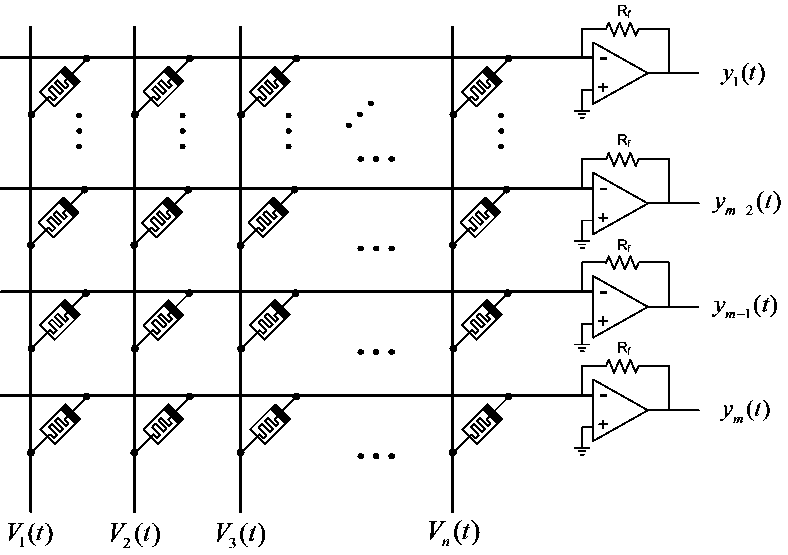}}
\caption{Typical neuromorphic computing system in which memristive
devices are used as a synapse.}
\label{figneuro} 
\end{figure}

As stated before, now it is well-known that the main contribution of
memristive devices in neuromorphic systems comes from their ability
to emulate the role of synapses in biological systems. This is
because of the fact that the memristance or memductance of
memristive devices can be tuned by the application of an appropriate
voltage or current signal to them. Therefore, it can be said that
memristance or memductance of memristive devices simply acts similar
to synaptic weights in biological systems. Figure \ref{figneuro}
shows a typical but not necessarily efficient neuromorphic system which is constructed based on these nanoscale passive elements and used in some recent studies \cite{Snider}. In this figure, memristive
devices are fabricated between two perpendicular wires in every
crosspoints of the crossbar and each rows of the crossbar is
connected to the virtually grounded input terminal of an operational
amplifier. Each of these operational amplifiers plays the role of
neurons in biological systems and their efficient hardware implementation is beyond the scope of this paper. Note that in this figure, memristive
devices are depicted explicitly to have better visibility.
Meanwhile, usage of the virtually grounded operational amplifiers in
this configuration is necessary to have a weighted sum of the
applied input voltages at their outputs. Since the combination of
each of these opamps with those memristive devices located on the
corresponding row of the crossbar creates a simple opamp-based
summing circuit, output of the {\it i}th neuron can be written as:
\begin{eqnarray}\label{eqa1}
y_i(t)=\sum_{j=1}^n\left(-\frac{R_{f}}{M_{i,j}(t)}\right)v_j(t),\quad
\forall i, \ 1\leq i\leq m
\end{eqnarray}
or
\begin{eqnarray}\label{eqa2}
y_i(t)=\sum_{j=1}^n\left(-R_fG_{i,j}(t)\right)v_j(t),\quad \forall i, \
1\leq i\leq m
\end{eqnarray}

where $M_{i,j}(t)$ and $G_{i,j}(t)$ are respectively the current
memristance and memductance of the memristive device at the
intersecting point of the {\it i}th horizontal and the {\it j}th
vertical wire of the crossbar, $R_{f}$ is a constant feedback resistor
of opamps, $m$ and $n$ are the number of rows and columns of the
crossbar respectively and $v_j$ is the voltage signal applied as an
input to the {\it j}th column of the crossbar. Note that although these two equations are the same (since $G_{i,j}(t)=\frac{1}{M_{i,j}(t)}$), but the first one should be used for the charge-controlled memristive devices while the second one is appropriate for the flux-controlled memrstive devices \cite{Chua}.

Here, we can see the first problem associated with this kind of
structures: synaptic weights ({\it i.e.}
$-\frac{R_{f}}{M_{i,j}(t)}$) relate inversely and nonlinearly to the
memristance of memristive devices. As a result, those learning
algorithms which are working by tuning of the memristance of
memristive devices may require an extra step to convert synaptic weights to the memristance of memristive devices and vice versa or at
least they should consider this note.

As another disadvantage, all synaptic weights
($-\frac{R_{f}}{M_{i,j}(t)}$ or $-R_fG_{i,j}(t)$) in the structure of Fig.
\ref{figneuro} are always negative (even they cannot become zero
although they may be sufficiently small). Adding auxiliary circuits
to the structure of Fig. \ref{figneuro} (for example by considering
two or more memristive devices to represent one synaptic weight) to
compensate this drawback will increase the complexity of the
associated learning method. One simple but effective solution to
overcome this problem for this kind of networks is to use those
learning rules that generate either positive or negative synaptic
weights. For example, in our recently proposed memristive
neuro-fuzzy computing system \cite{Neurofuzzy}, we have trained a
similar memristive network with a Hebbian-like learning rule with
only non-negative synaptic weights.

However, probably the most important disadvantage of using
memristive nanodevices in any kind of neuromorphic systems
independent from their structural configuration relates to
antithetical behavior of these passive elements during the learning
process. This antithetical behavior is the result of memristive
device's inability to act equally in the following two different
situations which occur frequently during the learning process: (i)
memristance increasing and (ii) memristance decreasing periods of
memristive devices. By precisely investigating the working procedure
of memristive devices while subjected to external voltages we can
observe that the response of these devices completely depends on the
polarity of the applied voltages and their initial conditions. To
clarify this issue better, here we  briefly review the behavioral
characteristics of some of the well-known memristive devices
fabricated recently.

Let's begin from the most primary fabricated memristive nanodevice,
{\it i.e.} HP device. In this device, a thin semiconductor
film of thickness $D$ is sandwiched between two metal contacts and
the boundary between doped (with length $w$) and undoped (with
length $D-w$) regions can be moved by the application of external
voltage bias $v(t)$. By considering a linear ionic drift in a
uniform field (ignoring boundary effects) with average ion mobility
$\mu_v$, memristance of this memristive device can be simply written
as \cite{williams}:
\begin{eqnarray}\label{eqa3a}
M(t)=R_{on}\frac{w(t)}{D}+R_{off}\left(1-\frac{w(t)}{D}\right)
\end{eqnarray}
\begin{eqnarray}\label{eqa3b}
\frac{\text{d}w(t)}{\text{d}t}=\mu_v\frac{R_{on}}{D}i(t)
\end{eqnarray}

where $R_{on}$ and $R_{off}$ are the minimum and the maximum
memristances that the memristive device can take and $i(t)=\frac{v(t)}{M(t)}$ is the
corresponding current passing through the device.

By solving equation \eqref{eqa3b} with respect to $w(t)$ we will have:
\begin{eqnarray}\label{eqa9}
w(t)&=&D-\sqrt{D^2-2\left[A+\mu_v\frac{R_{on}}{R_{off}}\int_{t_0}^t
v(t)\text{d}t\right]}\\
A&=&\left(Dw(t_0)-0.5w^2(t_0)\right)\nonumber
\end{eqnarray}

\begin{figure}[!t]
\centering 
{
\includegraphics[width=3.7in,height=2.2in]{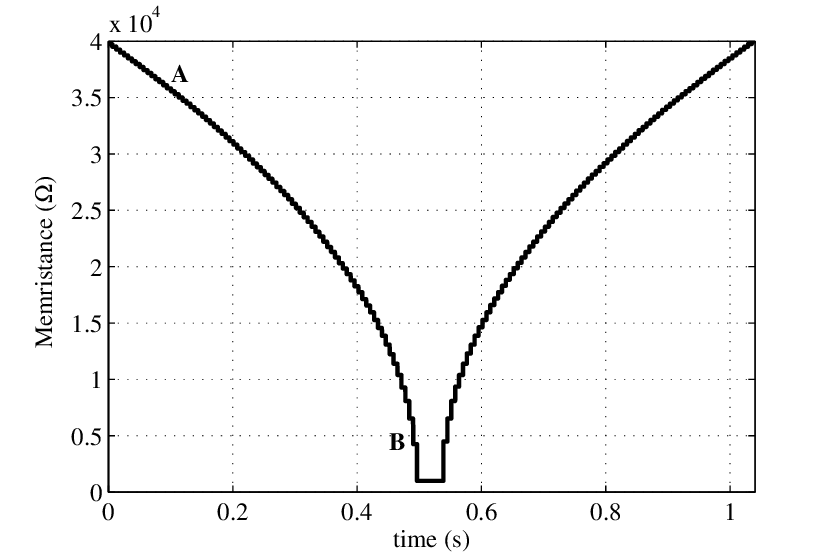}}
\caption{This figure shows how the memristance of this memristive
device varies (from $R_{off}=40k\Omega$ to $R_{on}=100\Omega$ and
vice versa) when the input is a successive positive voltage pulses
followed by a successive negative voltage pulses with the amplitude
of 2 volts each for a duration of 120$\mu s$ seconds. In this
simulation, $\mu_v$, $D$ and $M(t_0)$ are $10^{-13}$, $10^{-8}$ and
$40k\Omega$ respectively.}
\label{figHpsim} 
\end{figure}

Now, based on equations \eqref{eqa9} and \eqref{eqa3a} we can simply
draw the memristance of the HP device, {\it i.e.} $M(t)$, versus any
applied voltage for any initial condition ({\it i.e.} $w(t_0)$ or
equally $M(t_0)$). For example, Fig. \ref{figHpsim} shows how the
memristance of this memristive device varies (from $R_{off}$ to
$R_{on}$ and vice versa) when the input is a successive positive
voltage pulses followed by a successive negative voltage pulses with
the amplitude of 2 volts each for a duration of 120$\mu s$ seconds.
As this figure indicates, variation rate (slop) of the memristance
of this memristive device is proportional to the polarity of the applied
voltage and the initial memristance of the device. This means that the
memristance of the memristive device will change with different
amounts in different conditions even if the applied signal be the
same. In learning process, this will be translated to having
different and distinct time-varying learning rate for each synaptic
weight. For example, in Fig. \ref{figHpsim}, when the applied
voltage is positive, the learning rate is increasing with time while
by changing the polarity of the applied voltage, learning rate
begins to decrease with time. As a result, every memristive device used in any neuromorphic system
will have its own memristance-based learning rate. This may cause
the system to become unstable since the learning rate of some
synaptic weights may be very small (for example point A on Fig.
\ref{figHpsim}) while at the same time other's learning rate may be
very high (for example point B on Fig. \ref{figHpsim}). In the next
section, we will discuss a bit more about this fact ({\it i.e.}
possibility of the divergence of the memristive neuromorphic
system).

\begin{figure}[!t]
\centering \subfigure[]{
\label{fig:str:a} 
\includegraphics[width=3.1in,height=2.1in]{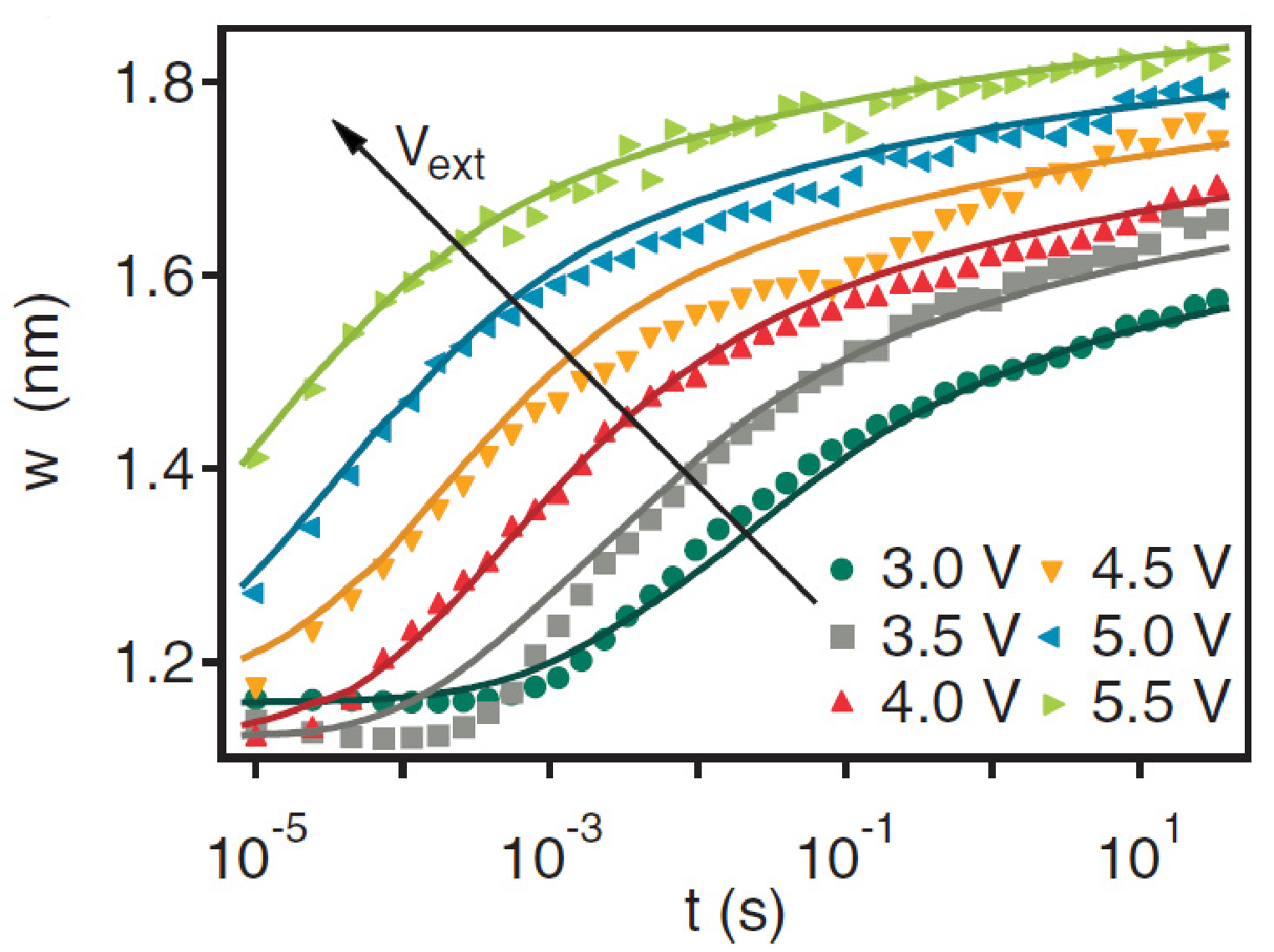}}
\vspace{0.05in} \subfigure[]{
\label{fig:str:b} 
\includegraphics[width=3.1in,height=2.1in]{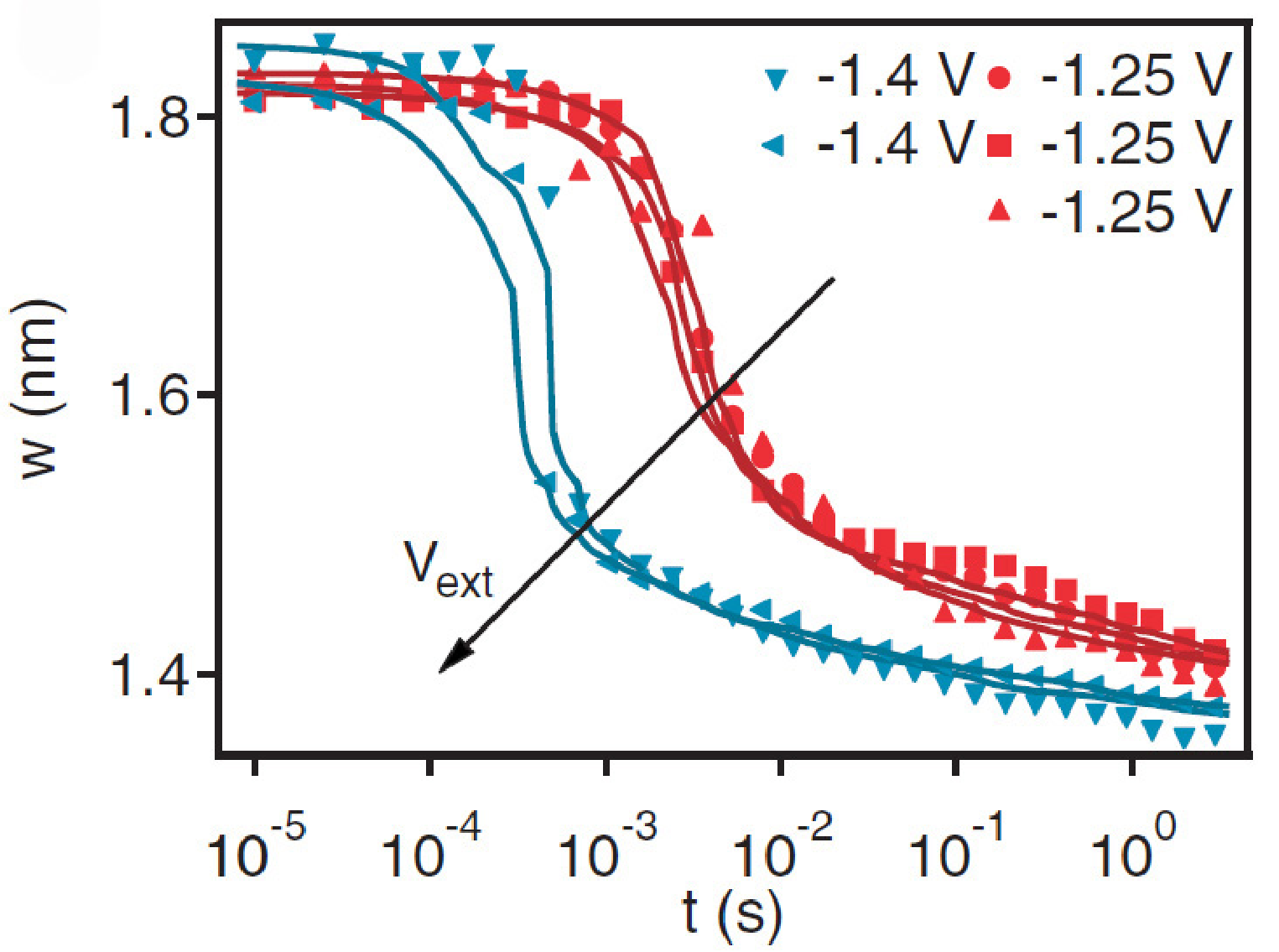}}
\caption{Experimental results of a dynamical testing protocol
applied to a set of Pt-$\text{TiO}_\text{2}$-Pt memristive devices
presented in \cite{strokuvJAP} (reprinted with permission from the
Journal of Applied Physics). Note that in these figures, $w$ is the
length of the undoped region of the device (unlike the case we had in HP device
which was the length of the doped region) and therefore higher $w$
will correspond to higher memristance. (a) This figure shows how the
state variable $w$ and consequently memristance of the device varies
during six off-switching tests with external voltage ranging from
3.0 to 5.5V. (b) In this figure we have the on-switching case which
is obtained by changing the polarity of the applied voltage. As can
be seen in these figures, variation rate of the memristance of the
memristive device varies in time and its variation rate at any time
depends completely to the memristance of the memristive device at
that moment. In addition, in figure (b) we can see the effect of the
explained positive feedback on the memristance of the memristive
device as well (abrupt change in the memristance of the device
compared with figure (a)).}
\label{fig:str} 
\end{figure}

This phenomenon (explained above) can be seen even in more realistic
memristive devices as well. For example, in \cite{strokuvJAP}, a
nonlinear memristive model of bipolar switching is presented which
is derived from the experimental results of a dynamical testing
protocol applied to a set of Pt-$\text{TiO}_\text{2}$-Pt crosspoint
devices. Here we have presented these experimental results in Fig.
\ref{fig:str} for convenience (note that in these figures, $w$ is
the length of the undoped region of the device (unlike the case we had in HP device which was the length of the doped region) and therefore
higher $w$ will correspond to higher memristance). Figure
\ref{fig:str:a} shows how the state variable $w$ and consequently
the memristance of the device varies during six off-switching tests with
external voltage ranging from 3.0 to 5.5V. By changing the polarity
of the applied voltage we will have the on-switching case which is
shown in Fig. \ref{fig:str:b}. As stated in \cite{strokuvJAP} as
well, it is clear from these figures that the switching speed
strongly depends on the current state of $w$ and the polarity of the
applied voltage. For example, Fig. \ref{fig:str:a} shows that
memristance of the device changes rapidly at first and its rate
begins to decrease by time. In Fig. \ref{fig:str:b}, we see a
completely different case: when the input voltage is negative,
memristance of the device changes more rapidly compared with the
case in which input voltage is positive. As stated before, this
means that based on the current state and condition, applied
identical positive or negative pulses may have completely different
effect on the memristance of the memristive device. In learning
algorithms like Spike Timing-Dependent Plasticity (STDP), this may
correspond to having different step size for each synaptic weight
which its characteristics depend not only on the current memristance
of the device but also to the polarity of the applied voltage.


\begin{figure}[!t]
\centering
\includegraphics[width=1.8in,height=1in]{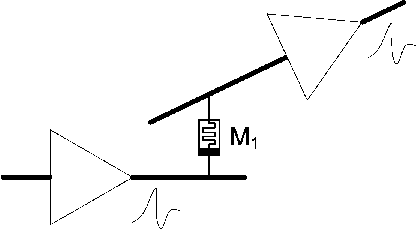}
\caption{Two neuron connected through a single memristive device.
The memristance of this memristive device should be decreased when
these two neurons fire Simultaneously.}
\label{fig:neuron} 
\end{figure}

To clarify the effect of this phenomenon on the working procedure of
memristive computing systems, consider the neuromorphic structure
shown in Fig. \ref{fig:neuron}. In this figure, two neurons are
connected to each other through the memristive synapse. It is
well-known that one of the main defining feature of connections
between neurons in biological systems is that they become stronger
when neurons fire together; hence the phrase ``neurons fire
together, wire together'', an event otherwise known as Hebbian
learning \cite{Hebb}. Various experiments have shown that this effect is most
pronounced early in the learning process when the increase in
connection strength is greatest while later learning merely
reinforces the links between neurons. However, that is somewhat at
odds with the actual behavior of memristive devices used as synapse.
For example, consider the memristive device connecting two neurons
in Fig. \ref{fig:neuron}. Memristance of this passive device
decreases when a voltage is applied to it which increases its
current which in return causes the memristance to drop further in a
kind of positive feedback effect. Note that although a lower
memristance allows more current to flow so this certainly increase
the strength of the connection as expected but because of the
mentioned positive feedback, later signals will have a bigger effect
on the connection than earlier ones. This is the opposite way round
to the way real neurons connect, where earlier signals have the
strongest effect. In other words, early experiments should have the
most impact on learning and not the later ones while this is not the
case in structures like the one shown in Fig. \ref{fig:neuron}. Here it is worth to mention that this problem cannot be solved by initializing the memristance of
memristive devices representing synaptic weights to rather small
values at the beginning of training phase as suggested in \cite{ramos}. This is because of the fact that in learning algorithms like STDP (as they
used), synaptic weights may change in both directions (up and down)
at any time and therefore initialization of the memristance of
memristive devices to rather small value is not effective and do not
solve the problem for subsequent steps of the learning process. Note
that this solution may be efficient only for those learning methods
in which synaptic weights vary only in one direction (for example
see \cite{IDS,Neurofuzzy}).

\begin{figure}[!t]
\centering 
{
\includegraphics[width=3.1in,height=2.1in]{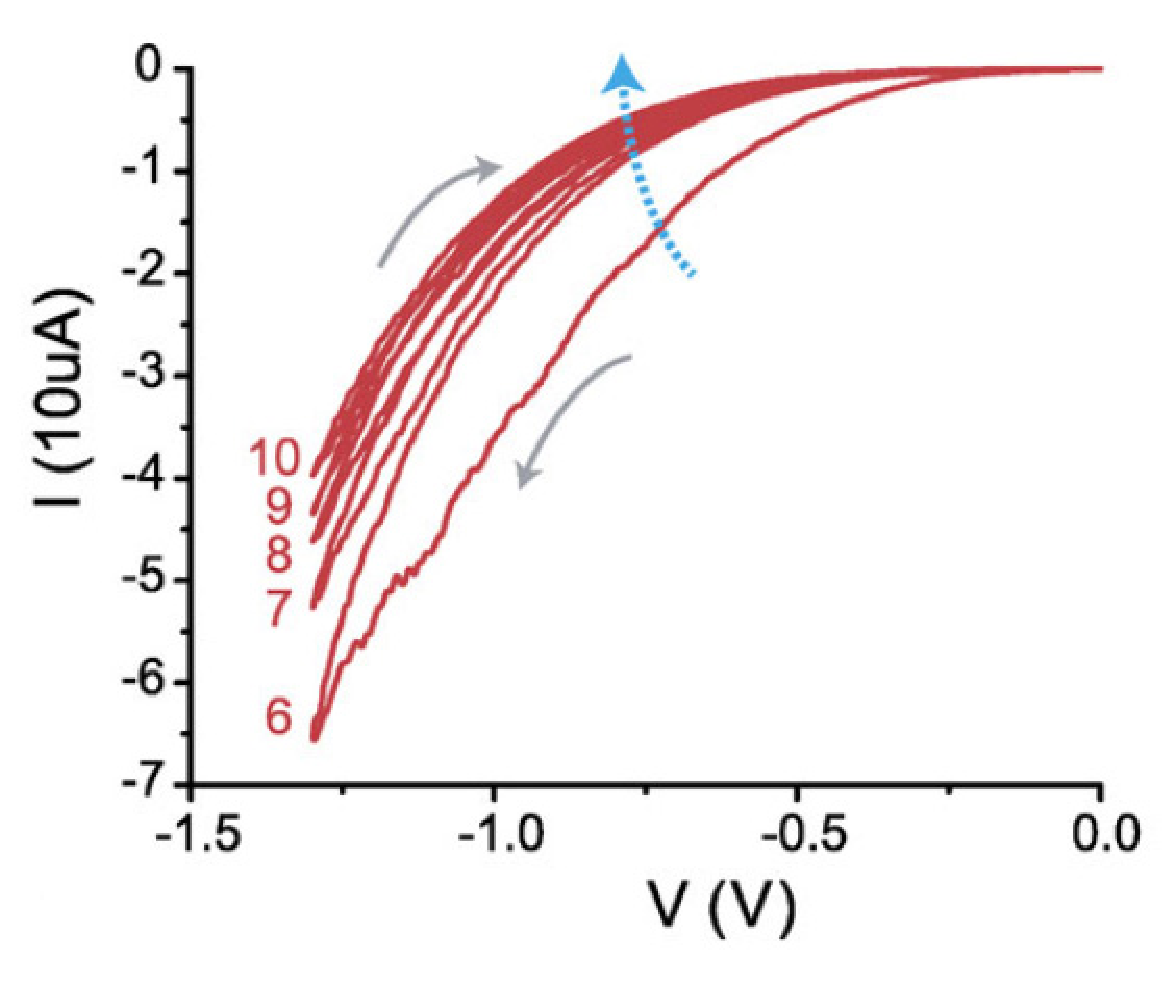}}
\caption{Typical resistance switching characteristics of the
flux-controlled Pd/WO3/W memristive device presented in
\cite{luspringer} (Reprinted with permission for Springer
Publishing). This figure shows the $I-V$ characteristic from five
consecutive negative voltage sweeps ($6-10$) showing a continuous
decrease in memductance. As can be seen in this figure, voltage
sweep number 6 has had more effect on the memristance of the
memristive device than for example voltage sweep number 10.}
\label{figlu} 
\end{figure}

This natural behavior of memristive devices and its associated
problem for neuromorphic systems as explained above can be seen in
some other reported memristive systems as well. For example, authors in \cite{luspringer} have presented a
new Pd/$\text{WO}_3$/W flux-controlled memristive devices. Although
they have argued that the memductance change in each voltage cycle
for their memristive device was roughly constant but their reported
experimental results show something else. For example, see Fig.
\ref{figlu} for the resistance switching characteristics of their
memristive device when subjected to five consecutive negative
voltage sweeps. From this figure of the paper, it is evident that
the voltage sweep no. 6 has changed the memductance of the device
more than what voltage sweep no. 10 or no. 9 has done. In addition,
those results (for example see Fig. 6(a) of the paper) which are
obtained by using their own reported final model for their
memristive device (although the model has this problem that by
setting the applied voltage into zero, memductance of the device
will continue changing) have also the same problem.

\begin{figure}[!t]
\centering 
{
\includegraphics[width=3.5in,height=2.3in]{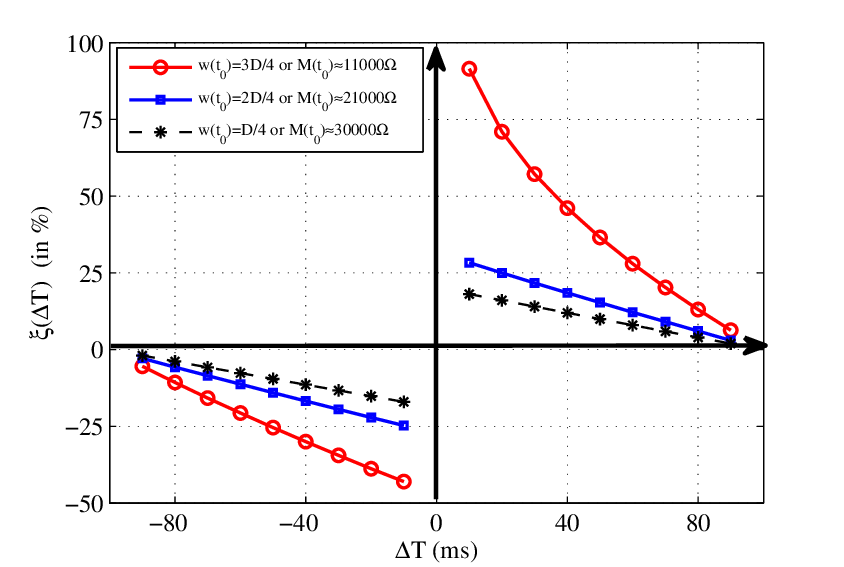}}
\caption{STDP update functions $\xi(\Delta T)$ for one memristive
synapse with three different initial memristances. It clearly shows
that the update function for each memristive synaptic weight depends
completely on its current memristance and therefore it will vary
with time during the learning process by the variation of the
memristance of the device.}
\label{updatefunction} 
\end{figure}

Finally, let's investigate the effect of this phenomenon on one of
the most famous learning algorithms of neuromorphic systems {\it
i.e.} STDP. The STDP algorithm is usually recognized by its update
function $\xi$ (changing amount of synaptic weight versus relative
timing of the spikes of neurons) \cite{Poo1,Poo2}. Dependency of the
memristive device's response on its initial memristance or
memductance results in having completely different STDP update
function for each initial condition. For example, in Fig.
\ref{updatefunction} we have plotted the update functions for one
memristive synapse (in this case HP device) for three different
initial conditions: $M(t_0)\approx11k\Omega$,
$M(t_0)\approx21k\Omega$ and $M(t_0)\approx30k\Omega$. It can be
inferred from this figure that during the learning phase of
memristive neuromorphic systems, each memristive device because of
its different memristance or memductance will have a completely
different update function $\xi$.

\section{The neuromorphic system realized through memristive
devices may not converge during learning phase}\label{Sec2}

\begin{figure}[!t]
\centering \subfigure[]{
\label{fig:costfuna} 
\includegraphics[width=3.5in,height=2.2in]{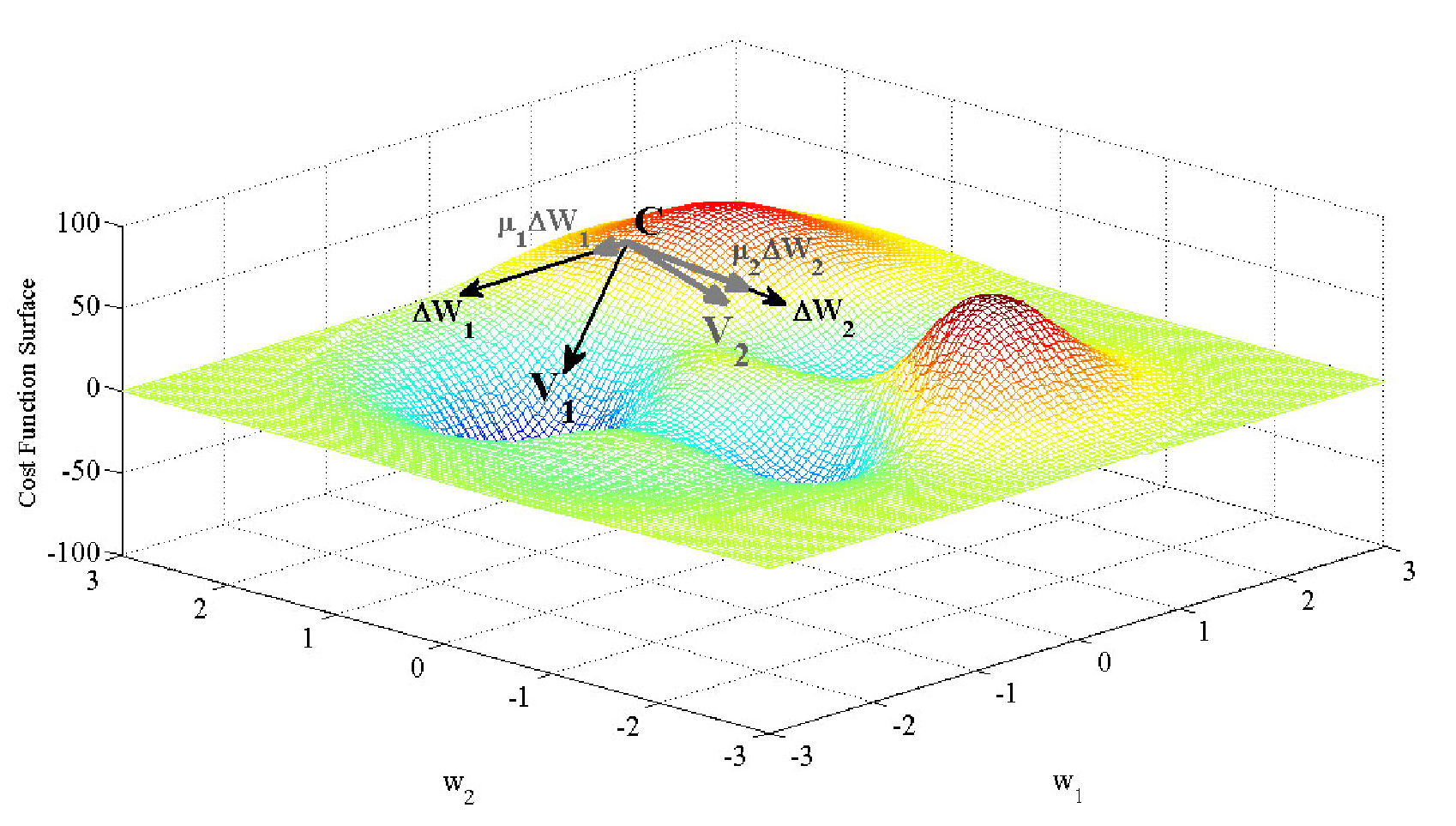}}
\subfigure[]{
\label{fig:costfunb} 
\includegraphics[width=3.5in,height=2.4in]{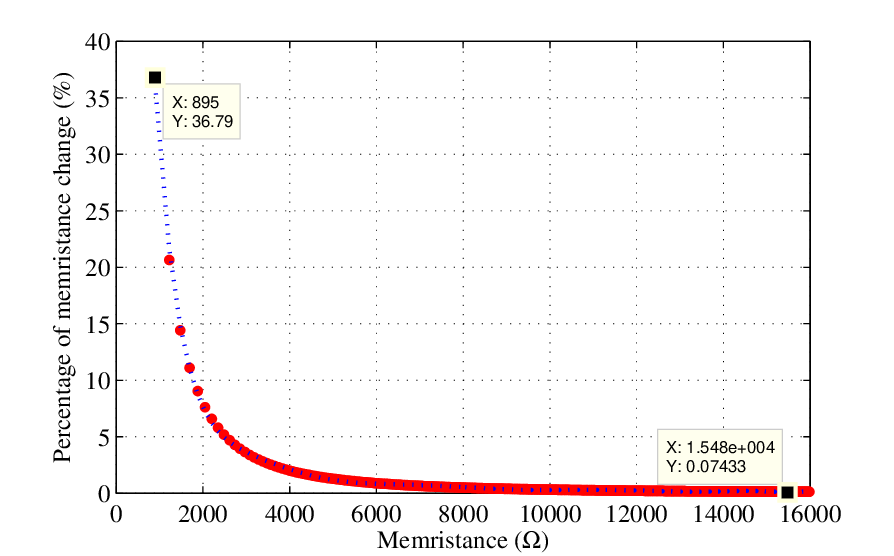}}
\caption{(a) Typical cost function of a neuromorphic system with two
synaptic weights. This figure shows that since memristance of
different memristive devices varies with different amounts due to
the applied voltage, the system may converge to a local minima
instead of global minima. (b) This figure shows how the memristance
of one memristive device with different initial memristances varies
due to the applied voltage. It shows that the ratio of memristance
change to initial memristance can differ significantly from one
memristive device to another memristive device which means that
those synaptic weights which are realize through this passive device
will have completely different learning rates.}
\label{fig:costfun} 
\end{figure}

In this section we will show why the memristive neuromorphic system
may become unstable during the learning phase or why it may not
converge to global extremum. For this purpose, assume that we have a
simple neuromorphic system with only two synaptic weights realized
through memristive devices. Moreover, let's assume that this system
has a cost function such as a typical one depicted in Fig.
\ref{fig:costfuna} which should be minimized with respect to
synaptic weights $w_1$ and $w_2$ during the learning phase. As can
be seen in Fig. \ref{fig:costfuna}, this typical cost function has
one global and two local minima. Now if we start from
point C on the surface of the cost function of Fig.
\ref{fig:costfuna}, then the application of an optimization method like
steepest descent will cause the weights vector $[w_1, w_2]$ to move
along vector $\overrightarrow{V_1}$ toward the global minimum. This
vector can be decomposed into two vectors $\overrightarrow{\Delta
w_1}$ and $\overrightarrow{\Delta w_2}$ which means that by
application of the optimization method, synaptic weights $w_1$ and
$w_2$ should move along vectors $\overrightarrow{\Delta w_1}$ and
$\overrightarrow{\Delta w_2}$ respectively. However, if the learning
rate of the synaptic weight $w_2$, {\it i.e.} $\mu_2$, be much higher
than the learning rate of the synaptic weight $w_1$, {\it i.e.} $\mu_1$,
the weight vector $[w_1, w_2]$ will move along vector $V_2$ toward
local minimum rather than global minimum. In this case, the
neuromorphic system under consideration will converge to a wrong
state. This problem shows itself better when we note that these
learning rates can differ significantly based on the current
memristance of those synaptic weights which are realized through
memristive devices. To verify this claim, we conducted a simple
experiment. In this experiment, we applied a positive voltage pulse
with fixed duration to a simple memristive device (using HP model)
with different initial memristances and measured the amount of
memristance change in percents due to the applied voltage. The result of this
experiment is presented in Fig. \ref{fig:costfunb}. This figure shows that if we implement a
neuromorphic system by using this memristive device, then those
synaptic weights which their corresponding memristive devices have
low memristances will have a learning rate of about
$36.79/0.07433\approx494.95$ times higher than the learning rate of
those synaptic weights which their corresponding memristive devices
have high memristances.

Finally, by considering Fig. \ref{fig:costfuna}, it should be noted
that this aforementioned problem about the possibility of the
divergence of the neuromorphic system can be relaxed by decreasing
the overall learning rate of the system for example by decreasing
the duration of the voltage pulse used for changing the memristance
of memristive devices. However, this solution will definitely
decrease the learning speed of the system.

\section{One simple solution to overcome some of these mentioned
problems}\label{Sec3}

\begin{figure*}[!t]
\centering \subfigure[]{
\label{fig:twomem:a} 
\includegraphics[width=0.8in,height=1in]{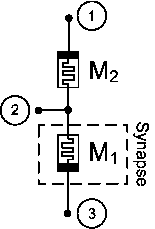}}
\subfigure[]{
\label{fig:twomem:a2} 
\includegraphics[width=2.2in,height=1.7in]{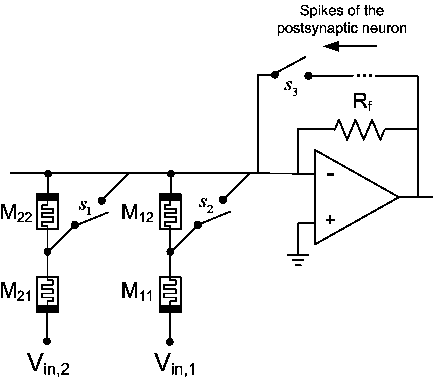}}
\subfigure[]{
\label{fig:twomem:b} 
\includegraphics[width=3.4in,height=1.9in]{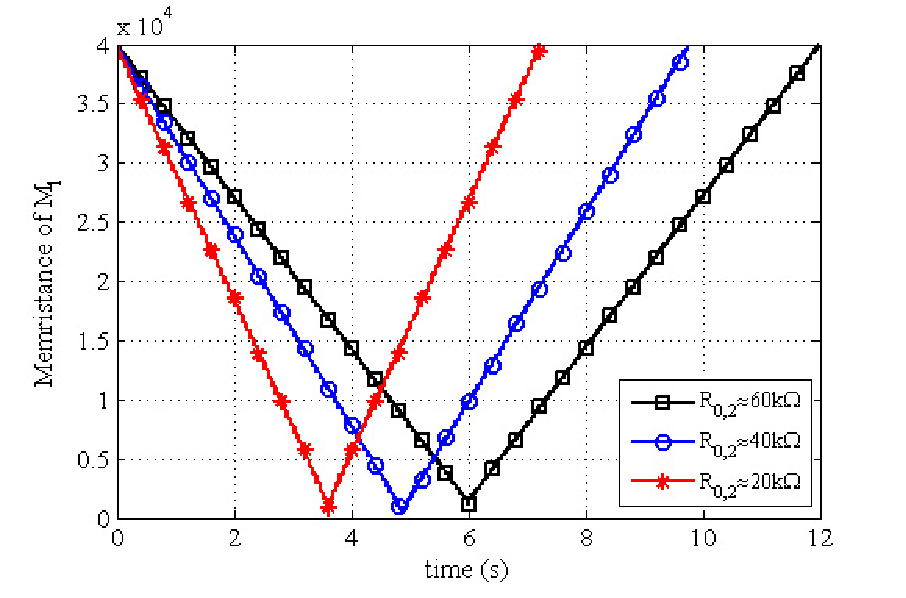}}
\subfigure[]{
\label{fig:twomem:c} 
\includegraphics[width=3.4in,height=1.9in]{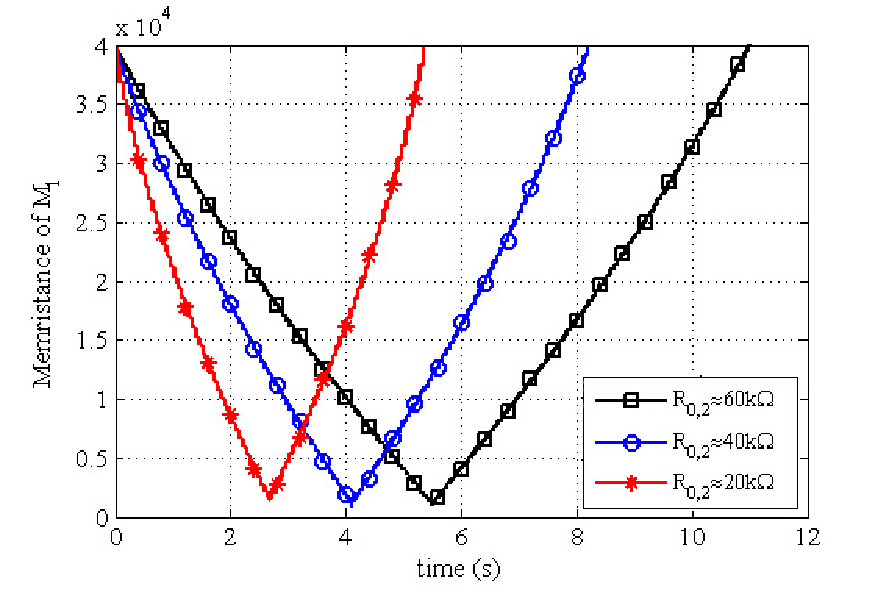}}
\subfigure[]{
\label{fig:twomem:d} 
\includegraphics[width=3.4in,height=1.9in]{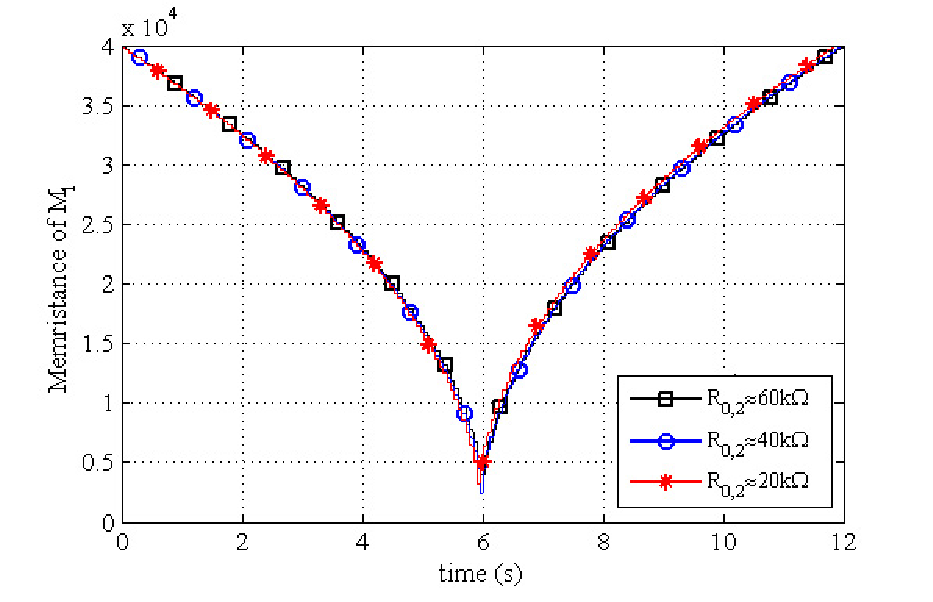}}
\caption{(a) Our proposed combinational element to use as a synapse
which consists of two serially connected memristive devices with
different polarities. Memristive device $M_1$ has the role of
synaptic weight and memristive device $M_2$ is an auxiliary device
added to solve some of the problems explained in this paper; (b)
This figure shows how our proposed memristive synapses can be used
in a neuromorphic system. (c) This figure shows how the memristance
of the memristive device $M_1$ in figure (a) varies (from
$R_{off,1}/2=40k\Omega$ to $R_{on,1}=100\Omega$ and vice versa) when
$\alpha=\Delta R_1/\Delta R_2=1$ and the input is a successive
positive voltage pulses followed by a successive negative voltage
pulses applied between input terminals 1 and 3. (d) The same as
figure (c) but by this difference that here $\alpha=\Delta
R_1/\Delta R_2=1/50\ll1$. (e) The same as figure (c) but by this
difference that here $\alpha=\Delta R_1/\Delta R_2=50\gg1$.}
\label{fig:twomem} 
\end{figure*}

To overcomes some of these problems, we propose to use the
combinational element shown in Fig. \ref{fig:twomem:a} as a synapse.
In this figure, two memristive devices with different polarities are
connected in series. In fact, memristive device $M_1$ plays the role
of synaptic weight and it will be used during normal working of the
system. On the other hand, memristive device $M_2$ is added only to
remove the aforementioned problem during learning phase. Figure
\ref{fig:twomem:a2} is a typical circuit which shows how these
combinational elements can be used to create a neuromorphic
computing system. In this circuit, devices $M_{11}$ and $M_{21}$ are
memristive synaptic weights and $V_{in,1}$ and $V_{in,2}$ are the
applied input voltages (either external voltages from sensory
neurons or internal voltages from other neurons in the system).
During the normal working phase of this neuromorphic system,
switches $s_1$ and $s_2$ will be close while switch $s_3$ is open
(note that one signal will be sufficient to control all of these
switches). In this case, the structure of Fig. \ref{fig:twomem:a2}
reduces to the ordinary memristive neuromorphic structures such as
the one depicted in Fig. \ref{figneuro} and the output of the system
will be the weighted sum of input signals. However, during the
learning phase, switches $s_1$ and $s_2$ become open while switch
$s_3$ is closed. In this configuration, a voltage difference between
spikes of presynaptic and postsynaptic neurons due to their timing
mismatch will drop this time across two memristive devices instead
of one compared with the structure of Fig. \ref{figneuro}. Now,
synaptic weights $M_1$ and $M_2$ see another memristive device with
opposite polarity on their way which may affect their memristance
variation rate during the learning phase.

Here, let's see how adding these auxiliary memristive devices
affects the updating process of memristive synaptic weights. For the
purpose of simplicity, we use the model of HP's memristive device
(considering a linear ionic drift in a uniform field and ignoring
boundary effects) for the elements in Fig. \ref{fig:twomem:a}. It
follows from the Kirchoff's laws that if two memristive devices
$M_1$ and $M_2$ which are connected in series have opposite
polarities, their total memristance can be written as \cite{yogesh}:
\begin{eqnarray}\label{eq:1}
\begin{split}
M_T(q)=\left(R_{0,1}-\eta\frac{\Delta
R_1q(t)}{Q_0}\right)+\left(R_{0,2}+\eta\frac{\Delta
R_2q(t)}{Q_0}\right)
\end{split}
\end{eqnarray}
which can be rewritten as:
\begin{eqnarray}\label{eq:2}
\begin{split}
M_T(q)=\left(R_{0,1}+R_{0,2}\right)-\eta\left(\Delta R_1-\Delta
R_2\right)\frac{q(t)}{Q_0}
\end{split}
\end{eqnarray}
where:
\begin{itemize}
\item $R_{0,i}$ for $i=1, 2$ is the effective memristance of the
memristive device $M_i$ at time $t_0$;
\item $\eta$ is the polarity of the memristive device which can be +1 and -1;
\item $\Delta R_i=R_{off,i}-R_{on,i}\simeq R_{off,i}$;
\item $q(t)$ is the amount of charge that has passed through the
devices;
\item $Q_{0,i}$ is the charge that is required to pass through the
memristive device $M_i$ for the dopant boundary to move through
distance $D$ where $D$ is the total length of the memristive device.
\end{itemize}

This equation shows that by connecting two memristive devices in
series but with different polarities we can suppress the q-dependent
component (second term in eq. \eqref{eq:2}). In addition, it is
clear that the overall behavior of these two memristive devices
depends on the relationship between $\Delta R_1$ and $\Delta R_2$ or
simply on $\alpha=\frac{\Delta R_1}{\Delta R_2}$. By properly
adjusting $\alpha$ we can choose how the memristance of the
memristive device $M_1$ changes versus the applied voltage. Here it
is worth to mention that two memristive devices connected in series
acts completely the same as a single memristive device (their
overall behavior is the same as the behavior of single memristive
device). However, it should be noted that here we have two
memristive devices during the learning phase but only one of them
will be used in the computation phase. In the other words, changing
rate of the memristance of the memristive device $M_1$ in Fig.
\ref{fig:twomem:a} can be completely different from the changing
rate of the overall memristance of two memristive devices which are
connected in series.

Figure \ref{fig:twomem:b} shows how the memristance of the
memristive device $M_1$ of Fig. \ref{fig:twomem:a} varies versus the
applied voltage when $\alpha=\frac{\Delta R_1}{\Delta R_2}=1$. This
simulation result is obtained by applying a successive positive
voltage pulses followed by a successive negative voltage pulses
(similar to what we had in Fig. \ref{figHpsim}) between input
terminals 1 and 3 and then plotting the memristance of $M_1$ versus
time. It can be inferred from this figure that when two serially
connected memristive devices have similar $\Delta R$s ({\it i.e.}
$\Delta R_1\approx\Delta R_2$), memristance of $M_1$ will change
almost linearly with time and therefore early and later applied
pulses will have the same effect on the memristance of the device.
In addition, it is interesting to note that as can be seen in Fig.
\ref{fig:twomem:b} as well, initial memristance of the memristive
device $M_2$ determines the variation rate of the memristance of
$M_1$. It means that during the learning phase of those neuromorphic
structures which are using this kind of synapse, we can change the
learning rate of the system simply by modifying memristance of
$M_2$. For example, to slow down the learning process we can simply
increase the memristance of $M_2$ which can be done even when the
system is running. In Fig. \ref{fig:twomem:c}, we have a completely
different case. This figure shows how the memristance of $M_1$
varies when $\alpha=\frac{\Delta R_1}{\Delta R_2}\ll1$ (in this
figure $\Delta R_2=50\Delta R_1$). It can be seen that when $\Delta
R_2$ is much larger than $\Delta R_1$, memristance of $M_1$ varies
with different rates depending on its memristance value. Again, by
changing the initial memristance of $M_2$, we can modify the
learning rate of the system. Finally, Fig. \ref{fig:twomem:d} shows
the case of $\alpha=\frac{\Delta R_1}{\Delta R_2}\gg1$ (in this
figure $\Delta R_1=50\Delta R_2$). Since the memristance of $M_2$ is
negligible compared with the memristance of $M_1$, memristive device
$M_2$ and consequently its initial memristance will have almost no
effect on the variation rate of the memristance of $M_1$ and
therefore we see the same result as we had in Fig. \ref{figHpsim}.
%
%

\begin{figure}[!t]
\centering 
{
\includegraphics[width=3.5in,height=2.3in]{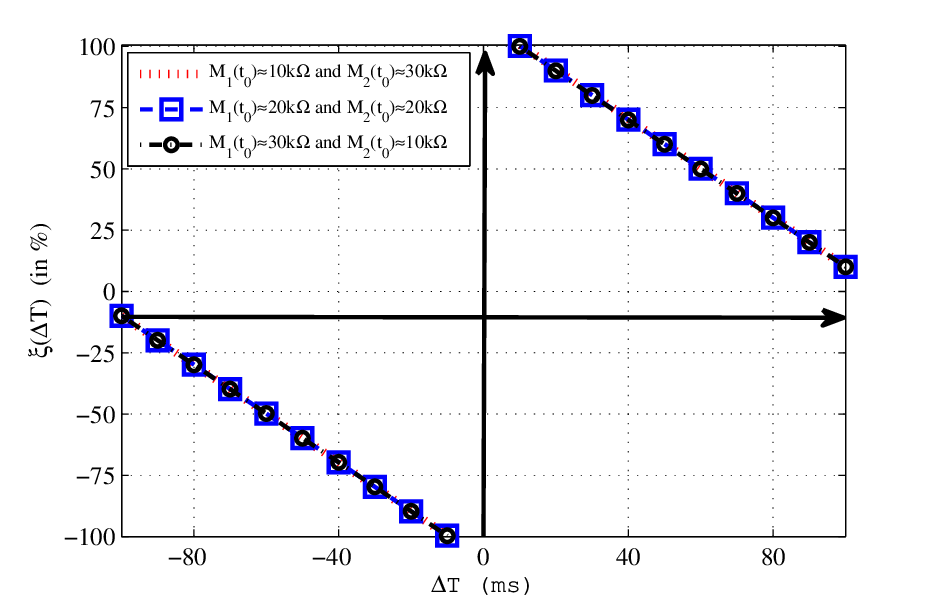}}
\caption{STDP update functions $\xi(\Delta T)$ of the memristive
device $M_1$ in Fig. \ref{fig:twomem:a} in three different
conditions (with different initial memristances) when $\Delta
R_1=\Delta R_2$ and the sum of initial memristances of two serially
connected memristive devices, {\it i.e.} $M_1$ and $M_2$, are equal
to the predetermined fix value (here $40k\Omega$). It can be
inferred from this figure that in our proposed configuration, the
STDP update function is almost the same for all memristive synapses
and does not depend on the initial memristance of memristive
device.}
\label{updatefunctiontwomem} 
\end{figure}

Another advantage that the structure of Fig. \ref{fig:twomem:a2}
offers is its ability to fix the problem of having different STDP
update function for each initial condition (memristance value) as we
explained before (see Fig. \ref{updatefunction}). This can be
achieved by satisfying two distinct conditions simultaneously: (i)
using memristive devices with the same $\Delta R$ in the entire
system (therefore having Fig. \ref{fig:twomem:b} for the variation
rate of all synaptic weights) and (ii) setting the initial
memristances of all pairs of memristive devices which are connected
in series in a way that the sum of their initial memristances be
equal to some predetermined fixed value. Note that this should be
done only once in the initialization phase. Since the two memristive
devices connected in series are similar but have different
polarities, the sum of their memristances at any time during the
learning phase will remain almost constant and equal to the
predetermined value. Therefore, during the initialization phase, it
is only required to set the memristance of each of the auxiliary
memristive device with the predetermined value minus the initial
memristance of its corresponding memristive device acting as a
synaptic weight. In this case, since every two connected memristive
devices are the same, as showed earlier in Fig. \ref{fig:twomem:b},
variation rate of the memristance of memristive devices will be
constant. In addition, since the total memristance of the two
serially connected memristive devices is always constant, the
current passing through the devices will always be the same and
therefore the initial memristance of the memristive device acting as
a synapse will not affect the STDP update function. Figure
\ref{updatefunctiontwomem} shows the STDP update function for this
special configuration for three different initial conditions. It can
be seen from this figure that unlike Fig. \ref{updatefunction} which
was obtained by using only one memristive device as synapse, the
STDP update function is almost the same for all synaptic weights
even with different initial memristances.

Finally, it is worth mentioning that although here we used the
simple HP model for our memristive devices, most of the problems and
solutions presented in this paper will be valid for other type of
memristive devices as well. For example, consider the boundary
effect. This effect will only show itself when the memristance of
the memristive device is near to its maximum or minimum value ({\it
i.e.} $R_{off}$ or $R_{on}$ respectively) and in other cases it will
be negligible.

%
%

\section{conclusion}\label{Sec4}
In this paper we showed that using a memristive device as a synapse
has some drawbacks. For example, we demonstrated that during the
learning process, variation rate of synaptic weights (memristance of
memristive devices) varies significantly from one memristive device
to another one based on the memristance of the device. This results
to have one distinct STDP update function per each individual
synaptic weight which may cause the neuromorphic system using this
learning method to diverge. To solve this problem, we proposed to
use two serially connected memristive devices with different
polarities as a synapses. Simulation results show that by proper
adjustment of the characteristics of these two memristive devices,
it is possible to overcome this aforementioned problem.

\end{document}